\documentclass[letterpaper, 10 pt, conference]{ieeeconf}

\usepackage{cite}
\usepackage{amsmath,amssymb,amsfonts}
\usepackage{graphicx}
\usepackage{hyperref} 
\usepackage{textcomp}
\usepackage{tabularx}
\usepackage{amsmath}
\usepackage{balance}
\usepackage{multirow}
\usepackage{dblfloatfix}
\usepackage{siunitx}
\usepackage{booktabs}
\usepackage{longtable}
\usepackage{cite}
\usepackage{color,colortbl}
\usepackage{algpseudocode}
\usepackage{algorithm}
\definecolor{Gray2}{gray}{0.9}
\definecolor{Gray}{gray}{0.7}

\usepackage{xcolor}
\usepackage[font=small]{caption}
\usepackage{titlesec}
\usepackage{subcaption}

\titlespacing*{\section}
{0pt}{1.2ex plus 0.5ex minus 0.2ex}{0.8ex}

\titlespacing*{\subsection}
{0pt}{1.0ex plus 0.5ex minus 0.2ex}{0.6ex}

\titlespacing*{\subsubsection}
{0pt}{0.8ex plus 0.3ex minus 0.1ex}{0.4ex}

\IEEEoverridecommandlockouts 

\begin{document}

\title{\LARGE \bf ImpedanceDiffusion: Diffusion-Based Global Path Planning for UAV Swarm Navigation with Generative Impedance Control
}
\author{Faryal Batool$^{*}$, Yasheerah Yaqoot$^{*}$, Muhammad Ahsan Mustafa, Roohan Ahmed Khan, \\ Aleksey Fedoseev, and Dzmitry Tsetserukou
\thanks{The authors are with the Intelligent Space Robotics Laboratory, Center for Digital Engineering, Skolkovo Institute of Science and Technology, Moscow, Russia. 
\tt \{faryal.batool, yasheerah.yaqoot, ahsan.mustafa, roohan.khan, aleksey.fedoseev, d.tsetserukou\}@skoltech.ru}
\thanks{*These authors contributed equally to this work.}
}

\maketitle

\begin{abstract}
Safe swarm navigation in cluttered indoor environment requires long-horizon planning, reactive obstacle avoidance, and adaptive compliance. We propose \textit{ImpedanceDiffusion}, a hierarchical framework that leverages image-conditioned diffusion-based global path planning with Artificial Potential Field (APF) tracking and semantic-aware variable impedance control for aerial drone swarms.

The diffusion model generates geometric global trajectories directly from RGB images without explicit map construction. These trajectories are tracked by an APF-based reactive layer, while a VLM--RAG module performs semantic obstacle classification with 90\% retrieval accuracy to adapt impedance parameters for mixed obstacle environments during execution.

Two diffusion planners are evaluated: (i) a top-view long-horizon planner using single-pass inference and (ii) a first-person-view (FPV) short-horizon planner deployed via a two-stage inference pipeline. Both planners achieve a 100\% trajectory generation rate across twenty static and dynamic experimental configurations and are validated via zero-shot sim-to-real deployment on Crazyflie~2.1 drones through the hierarchical APF--impedance control stack. The top-view planner produces smoother trajectories that yield conservative tracking speeds of 1.0--1.2\,m/s near hard obstacles and 0.6--1.0\,m/s near soft obstacles. In contrast, the FPV planner generates trajectories with greater local clearance and typically higher speeds, reaching 1.4--2.0\,m/s near hard obstacles and up to 1.6\,m/s near soft obstacles. Across 20 experimental configurations (100 total runs), the framework achieved a 92\% success rate while maintaining stable impedance-based formation control with bounded oscillations and no in-flight collisions, demonstrating reliable and adaptive swarm navigation in cluttered indoor environments.

Video of ImpedanceDiffusion: \href{https://youtu.be/zWotxL2fNjE}{https://youtu.be/zWotxL2fNjE}
\end{abstract}


\begin{figure}[t!]
\centering
\includegraphics[width=1\linewidth]{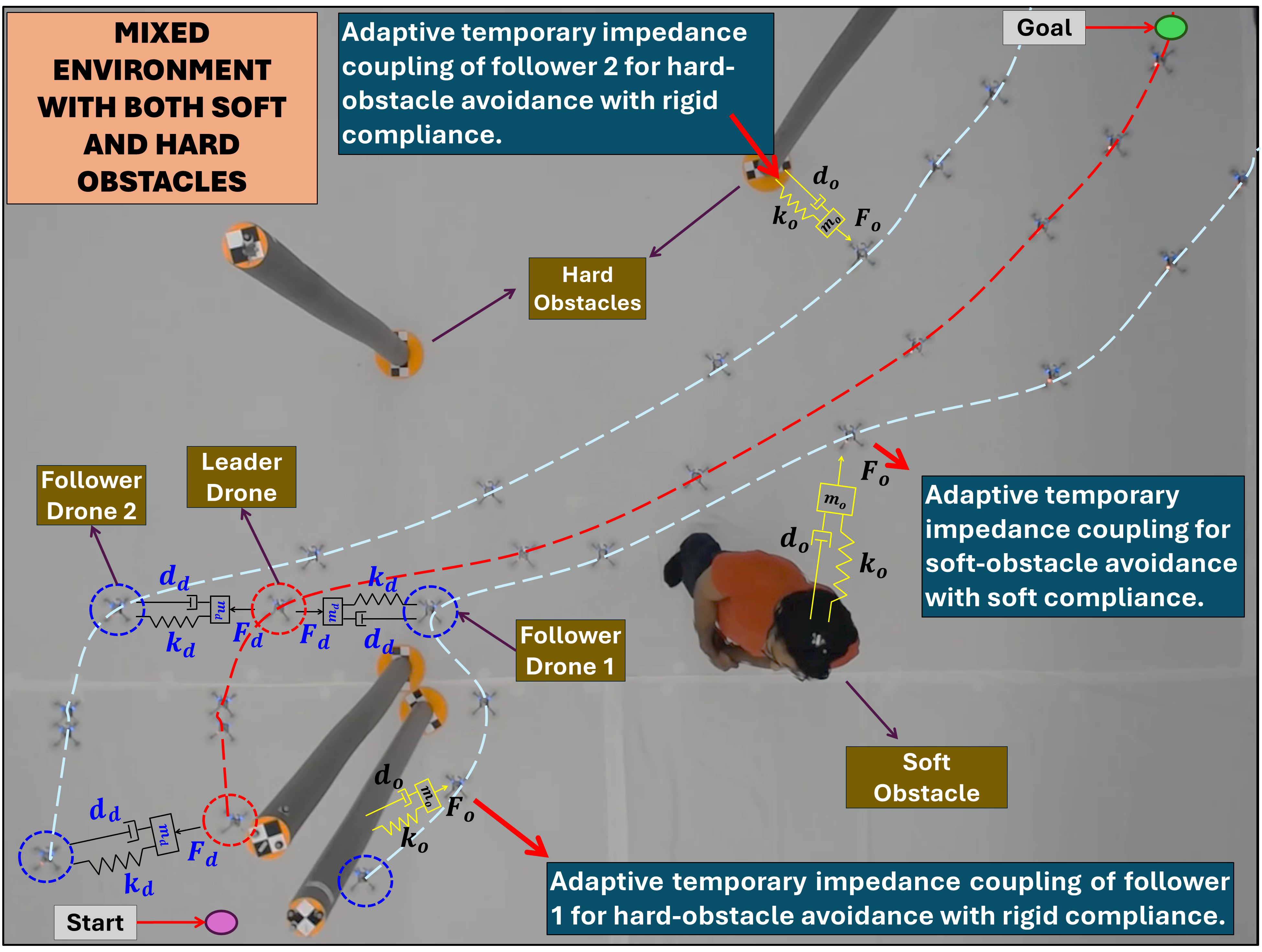}
\caption{\textit{ImpedanceDiffusion} in a mixed hard (rigid poles) and soft (human) obstacle environment. Diffusion-based planning generates the global path, APF ensures reactive avoidance (red dashed), and adaptive impedance links maintain formation cohesion and compliant interaction during flight.}
\label{fig:main_image}
\end{figure}

\section{Introduction}
Safe autonomous navigation in cluttered indoor environments remains a fundamental challenge for aerial robots. Corridors, warehouses, and laboratories present narrow passages, dynamic obstacles, varying illumination, and frequently lack reliable prior maps. Conventional navigation pipelines typically depend on explicit geometric representations (e.g., occupancy maps) and perception frameworks that fuse multiple sensors and intermediate outputs such as depth, segmentation, and mapping; these pipelines often degrade under limited onboard sensing and partial observability.

This motivates the following question: \emph{Can a drone infer safe global flight trajectories directly from a single RGB image, given only start and goal locations, without explicit maps or hand-crafted heuristics?}

Diffusion models offer a generative alternative for vision-conditioned trajectory synthesis. By learning trajectory distributions from image–trajectory pairs via iterative denoising, the model is trained to implicitly encode traversability cues from raw visual input, enabling smooth and coherent path generation without explicit geometric obstacle reconstruction. However, safe operation in human-shared environments requires more than geometric collision avoidance. It additionally demands adaptive compliance. Our prior work, \emph{ImpedanceGPT}~\cite{ImpedanceGPT}, introduced VLM--RAG-driven impedance adaptation for swarm navigation, but relied on Artificial Potential Fields (APF) for global planning and predefined impedance ranges per obstacle category.

In this work, we propose a hierarchical framework in which an image-conditioned diffusion model generates a \emph{global} path (replacing search-based global planning), while APF-based reactive tracking and generative impedance control provide robust avoidance for mixed obstacle environments (Fig.~\ref{fig:main_image}). We evaluate both a top-view diffusion planner for long-horizon global planning and an FPV-based diffusion planner for short-horizon iterative replanning, enabling a principled comparison between global planning and egocentric replanning strategies.

The main contributions of this work are:
\begin{itemize}
    \item An image-conditioned diffusion planner trained in simulation that generates smooth \emph{global} trajectories directly from a single RGB image and start/goal inputs, reducing reliance on classical search-based global planners.
    
    \item A hierarchical planning--control integration in which classical reactive tracking and adaptive impedance control collectively ensure stable trajectory following and obstacle avoidance of diffusion-generated global plans.

    \item A VLM--RAG framework that dynamically adapts swarm impedance parameters for mixed soft and hard obstacles.

    \item A quantitative comparison between top-view and FPV-conditioned diffusion planners, validated through zero-shot sim-to-real deployment on a Crazyflie swarm.
\end{itemize}

\section{Related Work}
Safe navigation in indoor aerial robotics requires the integration of global planning, local collision avoidance, and adaptive interaction control. While these components have been extensively studied individually, unified frameworks that combine visual trajectory generation, semantic reasoning, and compliant swarm coordination remain less explored within cohesive swarm-level architectures.

\subsection{Diffusion-Based Trajectory Generation}
Diffusion models have recently emerged as generative priors for robotic trajectory planning. Methods such as NoMaD, NavDP, and NaviDiffusor demonstrate goal-conditioned diffusion policies capable of generating smooth and collision-aware trajectories directly from visual observations \cite{NoMaD,NavDP,NaviDiffusor}. In mapless and long-horizon settings, DTG introduced a conditional diffusion architecture that optimizes travel distance and traversability \cite{DTG}. Similarly, DiPPeST applied image-conditioned diffusion planning to legged robots, achieving fast trajectory generation and reactive local refinement without classical search-based planners \cite{DiPPeST}. 

While diffusion-based planners address geometric feasibility at the global level, they lack mechanisms to regulate physical interaction once robots operate in close proximity to dynamic or heterogeneous obstacles. This limitation motivates the integration of active compliance control with learned global planning.

\subsection{Impedance Control for Aerial and Swarm Systems}
Impedance control, introduced by Hogan \cite{Paper_10}, establishes a desired force–motion relationship to enable compliant robot behavior during interaction. In aerial robotics, impedance-based approaches have improved trajectory tracking and disturbance handling, including Online Impedance Adaptive Control strategies \cite{Paper_03} and aerial manipulation frameworks \cite{Paper_06,Paper_07}. Intelligent impedance adaptations have also been explored to enhance robustness \cite{Paper_11}.

For swarm systems, virtual impedance links have been proposed to achieve safe and coordinated multi-quadrotor behavior \cite{Paper_12,Paper_13}. APF combined with impedance control has further improved swarm stability \cite{Paper_15}. However, most prior works rely on fixed or manually tuned impedance parameters that remain constant within a given scenario, and cannot explicitly differentiate interaction behavior across heterogeneous obstacle types. As a result, interaction compliance is typically scenario-dependent rather than obstacle-class-dependent.

\subsection{Vision-Language Models for Semantic Navigation}
Vision-Language Models (VLMs) provide semantic understanding beyond geometric perception. Recent aerial navigation works such as \emph{See, Point, Fly} demonstrate learning-free VLM-based waypoint grounding \cite{SPF}, while Transformer-based vision-language navigation frameworks enable quadcopter trajectory reasoning through multimodal fusion \cite{VLANaviQuad}. Other approaches integrate VLM reasoning for socially aware navigation and semantic trajectory selection \cite{Paper_17}, \cite{Paper_18}. Large multimodal models such as Molmo-7B-O, along with embedding methods such as Sentence-BERT further enhance contextual understanding \cite{Paper_19}, \cite{Paper_20}. However, VLM-based reasoning has not been directly coupled with adaptive impedance control for swarm-level compliant interaction.

While previous impedance-based navigation systems employ fixed parameter settings for predefined scenarios, real-world environments frequently contain mixed obstacle types that require differentiated interaction behavior. Building upon \cite{ImpedanceGPT}, which focused on fixed impedance parameters, the present work advances this direction by applying image-conditioned diffusion planning and introducing obstacle-class-dependent impedance adaptation. Compliance parameters are selected according to semantic scene understanding rather than global scenario categorization, enabling heterogeneous and mixed dynamic environments containing multiple obstacle types to induce distinct interaction behaviors within a single trajectory.

\section{ImpedanceDiffusion Framework}
\begin{figure*}[t]
    \centering
    \includegraphics[width=\textwidth]{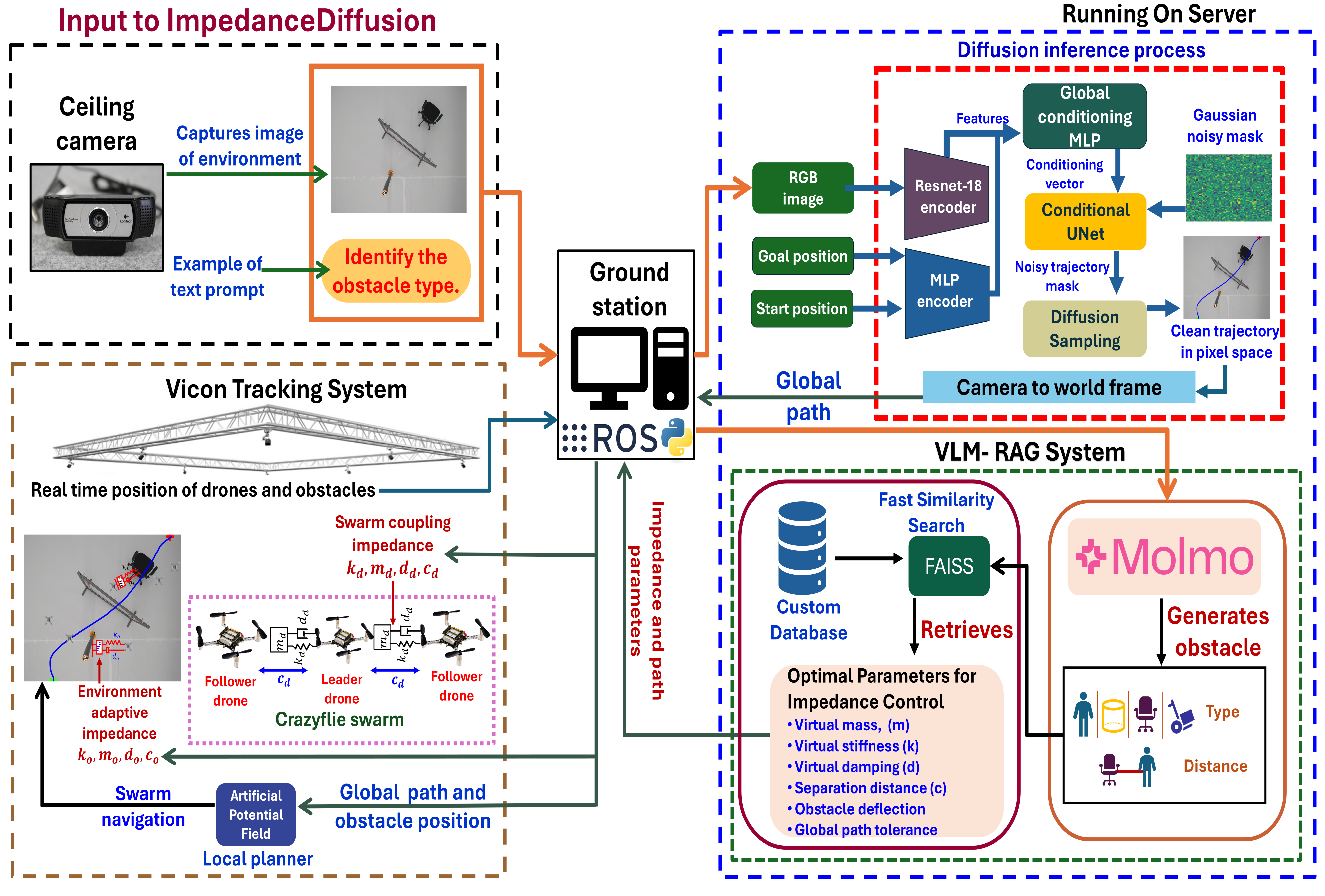}
    \caption{System architecture of ImpedanceDiffusion. A top-view image and user prompt are processed by Molmo to identify obstacle type. The VLM--RAG module retrieves the corresponding impedance parameters from a custom database. In parallel, the image is provided to the diffusion model to generate a global trajectory, which is executed using APF-based tracking with adaptive inter-drone and drone–obstacle impedance control.}
    \label{fig:system_architecture}
\end{figure*}

The proposed framework extends ImpedanceGPT~\cite{ImpedanceGPT} by employing a diffusion-based global trajectory generator within a hierarchical control architecture. The system comprises four modules: (1) an image-conditioned diffusion model for long-horizon global path planning, (2) an Artificial Potential Field (APF) as a local planner for tracking by the leader drone, (3) a VLM--RAG module for semantic scene understanding and impedance selection, and (4) a virtual impedance-based formation controller for follower drones as illustrated in Fig.~\ref{fig:system_architecture}. Conceptually, diffusion performs \emph{global plan synthesis}, whereas APF and impedance control form the \emph{execution layer}, responsible for tracking, short-horizon corrections, and maintaining safety margins during real flight.

\subsection{Semantic Obstacle Identification via VLM--RAG}
The VLM--RAG module follows the same semantic reasoning pipeline introduced in ImpedanceGPT~\cite{ImpedanceGPT}, with extensions to support heterogeneous environments. A Vision-Language Model (Molmo-7B-O \cite{molmo_quantized}) analyzes the top-view image to identify obstacle categories. The resulting semantic description is embedded and queried against a custom vector database using FAISS-based similarity search to retrieve impedance configurations suited to the detected scenario using Retrieval-Augmented Generation (RAG).


\subsection{Custom Database for Variable Impedance Scenarios}
To construct the impedance parameter database, 200 real-world swarm flight experiments were performed to collect interaction data in heterogeneous indoor environments containing both soft (humans) and rigid obstacles. For each obstacle type, multiple impedance configurations were evaluated to observe formation stability, oscillation behavior, jitter, and obstacle clearance performance. From these experiments, experimentally validated safe impedance parameter sets were identified for each obstacle category. The selected parameters correspond to configurations that minimized oscillations and jitter while maintaining stable, collision-free navigation across repeated trials.

The database stores the validated safe virtual mass ($m$), stiffness ($k$), damping ($d$), obstacle deflection limits, global path tolerance, and separation distance ($c$) for both drone--drone and drone--obstacle interactions. The resulting safety-consistent parameter sets are summarized in Table~\ref{tab:custom_database_single}. At runtime, the system retrieves the appropriate parameter set based on the detected obstacle class.
\begin{table}[t]
\centering
\footnotesize
\setlength{\tabcolsep}{3pt}
\renewcommand{\arraystretch}{1.1}
\caption{\textsc{Experimentally Validated Safe Virtual Impedance Link and Path Parameters for Drone--Drone and Drone--Obstacle Interactions, Stored in the Custom Database for Each Obstacle Type}}
\label{tab:custom_database_single}

\begin{tabular}{lccccc}
\hline
\textbf{Drone--Drone Parameters} & \textbf{Cyl.} & \textbf{Chair} & \textbf{Trolley} & \textbf{Gate} & \textbf{Human} \\
\hline

Virtual Mass (kg) & 1 & 1 & 0.8 & 1 & 5 \\
Virtual Stiffness (N/m) & 7 & 7 & 7 & 7 & 1 \\
Virtual Damping (Ns/m) & 3 & 3 & 3 & 3 & 2 \\

\hline
\textbf{Drone--Obstacle Parameters} & \textbf{Cyl.} & \textbf{Chair} & \textbf{Trolley} & \textbf{Gate} & \textbf{Human} \\
\hline

Virtual Mass (kg) & 1 & 0.8 & 0.8 & 1.2 & 1 \\
Virtual Stiffness (N/m) & 9 & 10 & 5 & 8 & 16 \\
Virtual Damping (Ns/m) & 5 & 5.5 & 3 & 5 & 4 \\

\hline
\textbf{Path Parameters} & \textbf{Cyl.} & \textbf{Chair} & \textbf{Trolley} & \textbf{Gate} & \textbf{Human} \\
\hline

Separation Distance (m) & 0.5 & 0.5 & 0.55 & 0.4 & 0.55 \\
Obstacle Deflection (m) & 0.65 & 0.8 & 1.2 & 0.45 & 1 \\
Global Path Tolerance (m) & 0.3 & 0.4 & 0.5 & 0.5 & 0.5 \\

\hline
\end{tabular}
\end{table}

\subsection{Diffusion-Based Global Planning}
The global planner is based on a conditional UNet-based diffusion model. The model predicts a pixel-space trajectory mask by iteratively denoising a noisy sample. The input is a three-channel mask  
\[
x_0 \in \mathbb{R}^{B \times 3 \times H \times W},
\]
where $B$ is the batch size, $H$ and $W$ are spatial dimensions, and the three channels encode the start pixel, goal pixel, and trajectory mask.

\paragraph{Forward Diffusion Process.}
Noise is progressively added to the clean mask using a squared-cosine schedule:
\begin{equation}
x_t = \sqrt{\overline{\alpha}_t}\, x_0 + \sqrt{1 - \overline{\alpha}_t}\, \epsilon,
\end{equation}
where $\alpha_t = 1 - \beta_t$ with $\beta_t$ the variance schedule,  
$\overline{\alpha}_t = \prod_{s=1}^{t} \alpha_s$ is the cumulative noise coefficient, and $\epsilon \sim \mathcal{N}(0,I)$.

\paragraph{Reverse Denoising Process.}
During inference, the conditional UNet predicts the clean mask $\hat{x}_0$, which is used in the Denoising Diffusion Probabilistic Model (DDPM) posterior:
\begin{equation}
x_{t-1} = \mu_t(x_t,\hat{x}_0) + \sigma_t z, \quad z \sim \mathcal{N}(0,I),
\end{equation}
where $\mu_t$ and $\sigma_t$ denote the posterior mean and variance. The denoising process iterates from $t=T$ to $t=0$, while start and goal channels are inpainted at each step to enforce boundary consistency. Here, $T$ denotes the total number of diffusion time steps.

\paragraph{Training Objective.}
The training objective combines trajectory reconstruction and endpoint accuracy:
\begin{equation}
\mathcal{L} = \lambda_{\text{path}}\,L_{\text{path}}
            + \lambda_{\text{endpoint}}\,L_{\text{endpoint}},
\end{equation}
where $\lambda_{\text{path}}$ and $\lambda_{\text{endpoint}}$ are the weights for trajectory reconstruction and endpoint accuracy respectively.
The trajectory reconstruction loss is
\begin{equation}
L_{\text{path}}
=
w_t \frac{1}{N}
\sum_{B,H,W}
\left(T^{\text{pred}}_{B,H,W} - T^{\text{gt}}_{B,H,W}\right)^2,
\end{equation}
where $T^{\text{pred}}$ and $T^{\text{gt}}$ are the predicted and ground-truth masks, respectively,  
$w_t$ is the trajectory-channel weight,  
and and $N = B \times H \times W$ is the total
number of pixels across the batch. The endpoint loss equals:
\begin{equation}
L_{\text{endpoint}} =
\frac{1}{2}\Big[
w_s \mathcal{L}_s + w_g \mathcal{L}_g
\Big],
\end{equation}
where $\mathcal{L}_s$ and $\mathcal{L}_g$ denote start and goal reconstruction losses with channel weights $w_s$ and $w_g$. The path term ensures geometric fidelity, while the endpoint term enforces boundary correctness.

\subsubsection{Training Pipeline and Dataset Generation}
We employ two diffusion-based global planners. The top-view planner is referred to as \textbf{Diffusion Planner 1} which is designed for long-horizon global planning. The FPV-based planner operates over a shorter horizon and is referred to as \textbf{Diffusion Planner 2}. This planner is an improved and extended version of the diffusion framework introduced in~\cite{humandiffusion}. These terms are used consistently throughout the paper.
Both \textbf{Diffusion Planner 1} (top-view) and \textbf{Diffusion Planner 2} (FPV) are trained using trajectories generated by an A* planner. The primary difference lies in the environments used for dataset generation. 

For \textbf{Diffusion Planner 1}, 10,000 A*-based trajectories are generated in the ProcTHOR environment~\cite{procthor_website}, with 1,500 validation and 1,647 test samples. 

For \textbf{Diffusion Planner 2}, 13,000 A*-based trajectories are generated in simulated indoor environments from~\cite{KennyLHW_VLNGo2Matterport_2024}, divided into 10,000 training, 1,500 validation, and 1,500 test samples.
 
All trajectories are restricted to traversable regions, implicitly enforcing collision-free behavior.
Across multiple configurations, training with 100 diffusion steps for 30 epochs achieved the best performance. The model is trained entirely in simulation and deployed in real-world experiments without fine-tuning, demonstrating zero-shot sim-to-real transfer.

\subsection{Local APF Tracking}
The diffusion model provides a global trajectory, while real-time tracking and reactive avoidance are handled by the Artificial Potential Field (APF) planner applied to the leader drone. At each control step, the next waypoint on the diffusion path is treated as a temporary goal. The total force acting on the leader is
\begin{equation}
F_{\text{total}} = F_{\text{attraction}} + F_{\text{repulsion}}.
\end{equation}

The attractive force toward the waypoint is
\begin{equation}
F_{\text{attraction}}(d_{\text{g}}) = k_{\text{att}} d_{\text{g}},
\end{equation}
where $d_{\text{g}}$ is the distance to the waypoint and $k_{\text{att}}$ is the attraction gain.

The repulsive force from nearby obstacles is defined as
\begin{equation}
{F}_{\text{rep}}(d_{\text{obs}}) =
\begin{cases}
\mathbf{0}, & d_{\text{obs}} > d_{\text{safe}}, \\[6pt]
k_{\text{rep}}
\left( \frac{1}{d_{\text{obs}}} - \frac{1}{d_{\text{safe}}} \right)
\frac{1}{d_{\text{obs}}^{2}},
& d_{\text{obs}} \le d_{\text{safe}},
\end{cases}
\end{equation}
where \(d_{\text{obs}} = \|\mathbf{x} - \mathbf{o}\|\) is the Euclidean distance between the drone position \(\mathbf{x}\) and the obstacle position \(\mathbf{o}\), \(d_{\text{safe}}\) is the safety radius, and \(k_{\text{rep}}\) is the repulsion gain. 

\subsection{Impedance Control for Swarm Coordination and Obstacle Interaction}
To achieve compliant swarm coordination and safe obstacle avoidance, two distinct impedance mechanisms are employed:
(i) a dynamic drone–drone formation impedance and 
(ii) a distance-based drone–obstacle interaction impedance.

\subsubsection{Drone–Drone Formation Impedance}
Each follower drone maintains a virtual impedance link with the leader. The leader velocity is computed as
\begin{equation}
\mathbf{v}_L(t) = \frac{\mathbf{x}_L(t) - \mathbf{x}_L(t-\Delta t)}{\Delta t},
\end{equation}
where $\mathbf{x}_L$ denotes the leader position and $\Delta t$ is the sampling interval.

The virtual interaction is modeled as a mass–spring–damper system. For each follower, the impedance dynamics are governed by
\begin{equation}
m_d \ddot{\mathbf{z}} + d_d \dot{\mathbf{z}} + k_d \mathbf{z} = {F}_{\text{ext}}(t),
\end{equation}
where:
\begin{itemize}
    \item $\mathbf{z}$ is the virtual formation displacement,
    \item $m_d$, $d_d$, $k_d$ are the virtual mass, damping, and stiffness parameters for drone--drone formation,
    \item and ${F}_{\text{ext}}(t)$ represents the virtual external force induced by the leader drone.
\end{itemize}

The final follower target position, in a drone formation is
\begin{equation}
\mathbf{x}_j =
\mathbf{x}_L
+ \beta \mathbf{z}
+ \mathbf{r}_j,
\end{equation}
where $\beta$ is a scaling factor and $\mathbf{r}_j$ is the geometric formation offset defined as:
\begin{equation}
\mathbf{r}_j =
\begin{bmatrix}
R \cos\theta_j \\
R \sin\theta_j \\
0
\end{bmatrix},
\end{equation}
here $R$ is the separation distance and $\theta_j$ defines the angular distribution of followers.

The parameters $(m_d, d_d, k_d)$ are adaptively modulated based on proximity to soft obstacles (humans), using hysteresis thresholds:
\begin{equation}
\text{NearHuman}_j =
\begin{cases}
1, & d_j \le d_{\text{enter}} \\
0, & d_j \ge d_{\text{exit}}
\end{cases},
\end{equation}
where $d_j$ is the minimum distance between follower $j$ and any human obstacle. When $d_j > d_{\text{exit}}$, the inter-drone impedance parameters revert to their default values corresponding to hard-obstacle interaction.

\subsubsection{Drone--Obstacle Interaction Impedance}
When a follower enters a predefined deflection radius around the nearest obstacle, a temporary impedance interaction is activated. Let $\mathbf{x}_j$ be the follower position and $\mathbf{x}_o$ the obstacle position. The relative vector and distance are defined as:
\begin{equation}
\mathbf{r} = \mathbf{x}_j - \mathbf{x}_o, 
\qquad
d = \|\mathbf{r}\|.
\label{eq:relative_distance}
\end{equation}

If $d < d_{\mathrm{def}}$, where $d_{\mathrm{def}}$ is the deflection distance, the penetration depth is defined as:
\begin{equation}
\delta = d_{\mathrm{def}} - d.
\label{eq:penetration_depth}
\end{equation}

To incorporate impedance behavior, the temporal variation of the penetration depth is computed in discrete time. The penetration velocity and acceleration are approximated as:
\begin{equation}
\dot{\delta}_t = \frac{\delta_t - \delta_{t-1}}{\Delta t},
\qquad
\ddot{\delta}_t = \frac{\dot{\delta}_t - \dot{\delta}_{t-1}}{\Delta t},
\end{equation}
where $\Delta t$ denotes the control timestep. These quantities represent the rate at which the follower enters or exits the obstacle interaction region.

The normal direction of interaction is defined as
\begin{equation}
\hat{\mathbf{n}} = \frac{\mathbf{r}}{\|\mathbf{r}\|}.
\end{equation}

A virtual impedance force along the obstacle normal is then computed using a mass--spring--damper model:
\begin{equation}
F_n = k_o \delta + d_o \dot{\delta} + m_o \ddot{\delta},
\end{equation}
where $k_o$, $d_o$, and $m_o$ are obstacle-dependent virtual stiffness, damping, and mass parameters, respectively. In practice, the acceleration term may have limited influence due to small penetration variations; however, it is retained for completeness of the impedance formulation.

The parameters $k_o$, $d_o$, and $m_o$ are selected according to the detected obstacle class from the database. Soft obstacles use lower stiffness and higher damping for compliant deflection, while rigid obstacles use higher stiffness for stronger repulsion. Gate structures share grouped parameters to maintain coordinated passage.

The obstacle-induced displacement is obtained by converting the virtual force into a position correction through discrete-time integration. Using a constant-acceleration approximation, the induced displacement is given by
\begin{equation}
\mathbf{u}_{\mathrm{obs}} = \frac{1}{2} \frac{F_n}{m_o} \Delta t^2 \, \hat{\mathbf{n}}.
\label{eq:obs_displacement}
\end{equation}

The updated follower position becomes
\begin{equation}
\mathbf{x}_j^{\mathrm{new}} = \mathbf{x}_j + \mathbf{u}_{\mathrm{obs}}.
\label{eq:updated_position}
\end{equation}

Unlike formation impedance, the obstacle interaction is implemented in discrete time and avoids continuous-time ODE integration.

\section{Experimental Setup}
A total of 20 experimental configurations were evaluated across eight distinct scenario types to assess the proposed \textit{ImpedanceDiffusion} framework in terms of global trajectory generation, semantic understanding, and variable impedance-based swarm navigation. The experiments covered static and dynamic scenarios, as well as sparse and cluttered environments, to assess robustness under varying complexity levels. Top-down images used for semantic reasoning were captured under consistent daylight conditions to ensure reliable visual perception. Multi-colored obstacles were introduced to evaluate the semantic classification capability of the VLM and mitigate potential color bias.
Two diffusion models were evaluated. The first model was trained on top-view images and its inference was executed once per trajectory. The second model was trained on first-person-view (FPV) data and its inference was done twice to generate a complete trajectory: first, from the start to an intermediate waypoint, and then to the final goal.

The VLM--RAG and diffusion models were executed on a high-performance workstation equipped with an NVIDIA RTX 4090 GPU (24GB VRAM) and an Intel Core i9-13900K processor. Due to memory constraints, the quantized Molmo-7B-O BnB 4-bit model was employed for semantic reasoning. For real-world validation, Crazyflie 2.1 drones were used to implement swarm navigation in physical indoor environments \cite{crazyflie}. The VLM--RAG module received the following structured prompt for obstacle detection:
\textit{``Analyze the drone arena image and identify all the obstacles present on the white background of the image.''}

\section{Experimental Results}
All 20 experiments were conducted five times across eight distinct scenarios, including static and dynamic environments with hard, soft, and mixed obstacle configurations. These repeated trials were performed to evaluate the success rate of the proposed framework. Across the resulting 100 runs, the framework achieved a \textbf{92\%} success rate, with failures primarily attributed to hardware limitations or communication loss. No failures were attributed to trajectory generation or impedance control instability. In this paper, we present eight representative experiments to illustrate the characteristic behaviors of the proposed framework under different environmental conditions. 

\subsection{VLM--RAG Retrieval}
Across the 20 experimental configurations in 8 scenario types, the VLM--RAG module was evaluated on its ability to correctly identify the object type and retrieve the corresponding impedance parameters. Out of the 20 cases, 18 were correctly classified and matched with the appropriate parameter set, resulting in a retrieval accuracy of \textbf{90\%}.


\subsection{Swarm Navigation in Static Environment}

\subsubsection{Experiment 1: Hard Obstacles with Gate}
\begin{figure}[htbp]
    \centering
    \includegraphics[width=\linewidth]{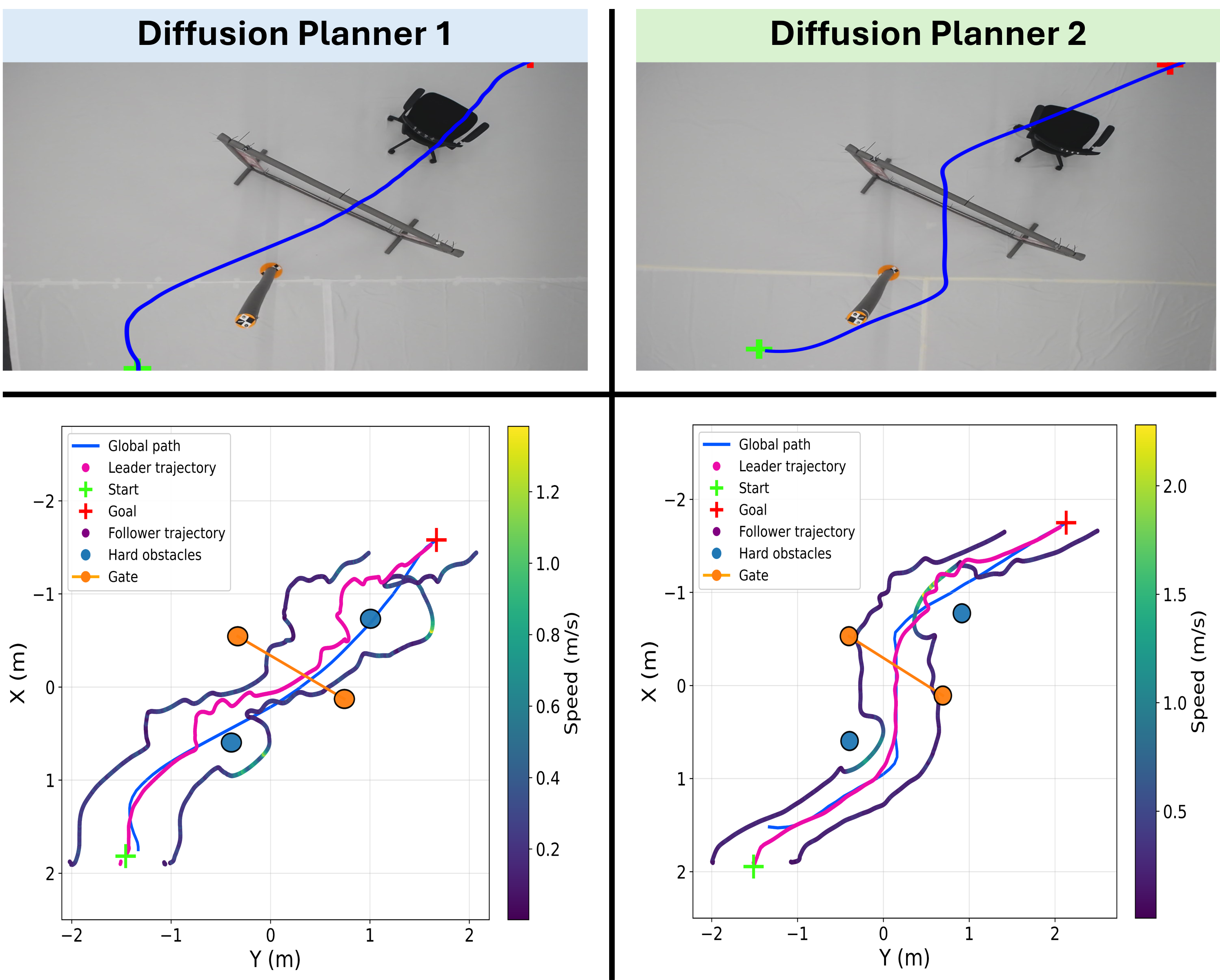}
    \caption{Static environment with hard obstacles and a gate. Rigid compliance with small deflection is observed near hard objects.}
    \label{fig:static_exp_01}
\end{figure}

Fig.~\ref{fig:static_exp_01} presents a static environment containing a cylindrical obstacle, a chair, and a gate. Both diffusion planners are provided with identical top view images, start, and goal inputs. Diffusion Planner 1 functions as a global image-conditioned planner, while Diffusion Planner 2 operates as an FPV-based local planner, generating trajectories conditioned on the drone’s forward-facing view. Consequently, Planner 2 adapts more explicitly to the locally visible obstacle geometry.
Diffusion Planner 1 generates smoother and more direct trajectories. In contrast, Diffusion Planner 2 maintains greater clearance near obstacle boundaries, as reflected in the velocity-colored plots. When navigating near hard obstacles, the follower speed under Planner 1 remains between 1.0--1.2 m/s. Planner 2 maintains slightly higher speeds of 1.5--2.0 m/s due to increased local clearance, which reduces APF-induced repulsive forces and minimizes corrective accelerations.

For hard obstacles, high stiffness and moderate damping produce small deflections and reduced compliance. This maintains tight leader–follower coupling, enabling obstacle avoidance with minimal lateral deviation. Minor oscillations arise when the follower momentarily advances ahead of the leader and is pulled back by the impedance link, appearing as small jitters in the plots.

\subsubsection{Experiment 2: Cluttered Scene with Hard and Soft Obstacles}
\begin{figure}[htbp]
    \centering
    \includegraphics[width=\linewidth]{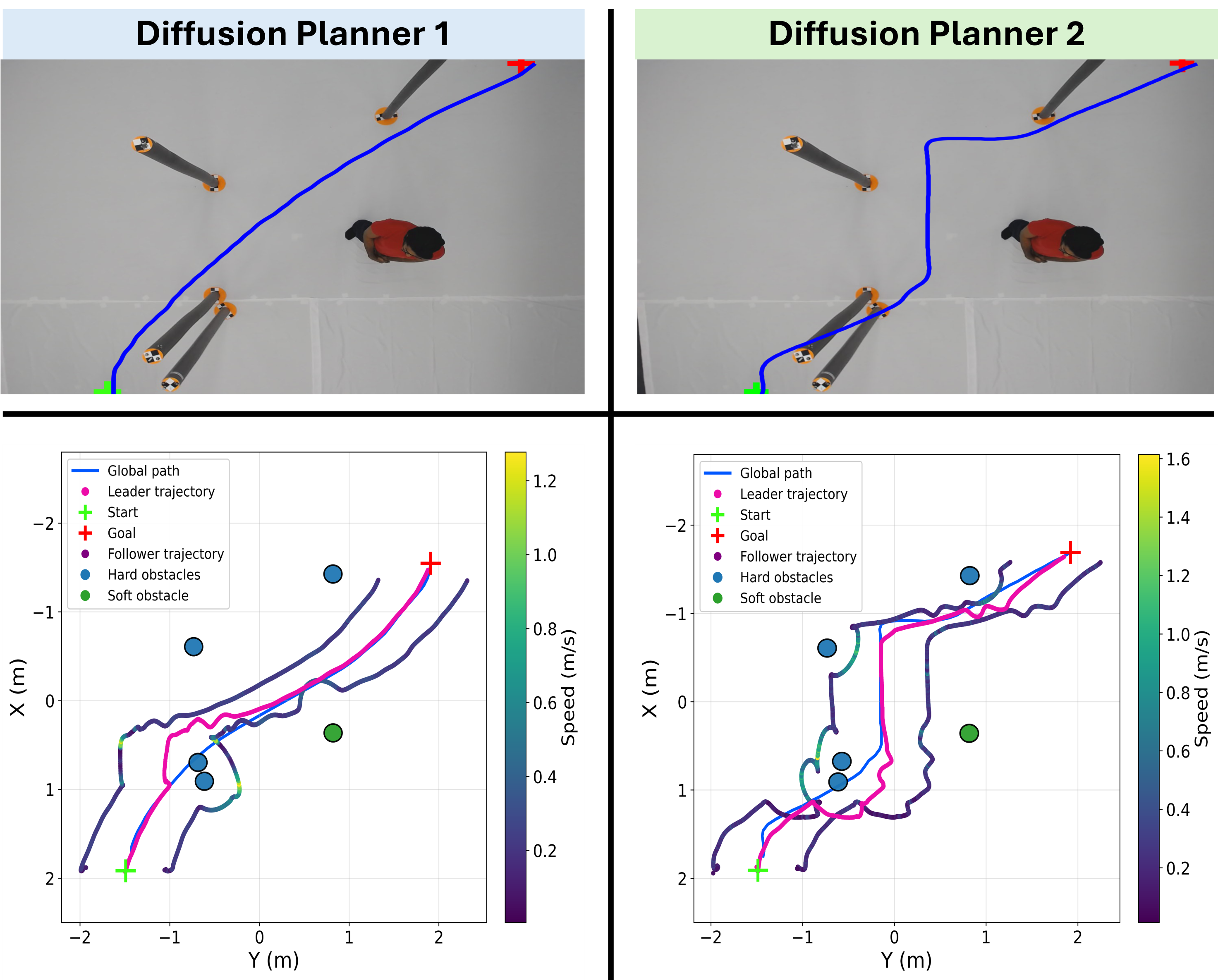}
    \caption{Static cluttered environment with hard obstacles and one human. Soft compliance near the human results in larger deflection and reduced speed.}
    \label{fig:static_exp_02}
\end{figure}

Fig.~\ref{fig:static_exp_02} illustrates a cluttered static environment containing four cylindrical obstacles and one human. Around hard obstacles, both planners exhibit rigid compliance characterized by small deflection distances and relatively high speeds, approximately 1.0 m/s for Planner 1 and 1.6 m/s for Planner 2.
In contrast, the response near the soft obstacle (human) differs significantly. The follower speed decreases to approximately 0.6 m/s for both planners, and the deflection distance increases compared to hard obstacles. This behavior aligns with the impedance switching strategy proposed in ImpedanceGPT~\cite{ImpedanceGPT}, where humans are assigned lower stiffness and moderate damping. In our case, these parameters are slightly higher due to the use of variable impedance control.

Planner 2 consistently maintains higher speeds overall, as its FPV-based local diffusion planning provides greater clearance in directly visible regions, thereby reducing APF-induced corrective forces compared to Planner 1.

\subsection{Swarm Navigation in Dynamic Environment}

\subsubsection{Experiment 3: Single Dynamic Human with Hard Obstacle}
\begin{figure}[htbp]
    \centering
    \includegraphics[width=\columnwidth]{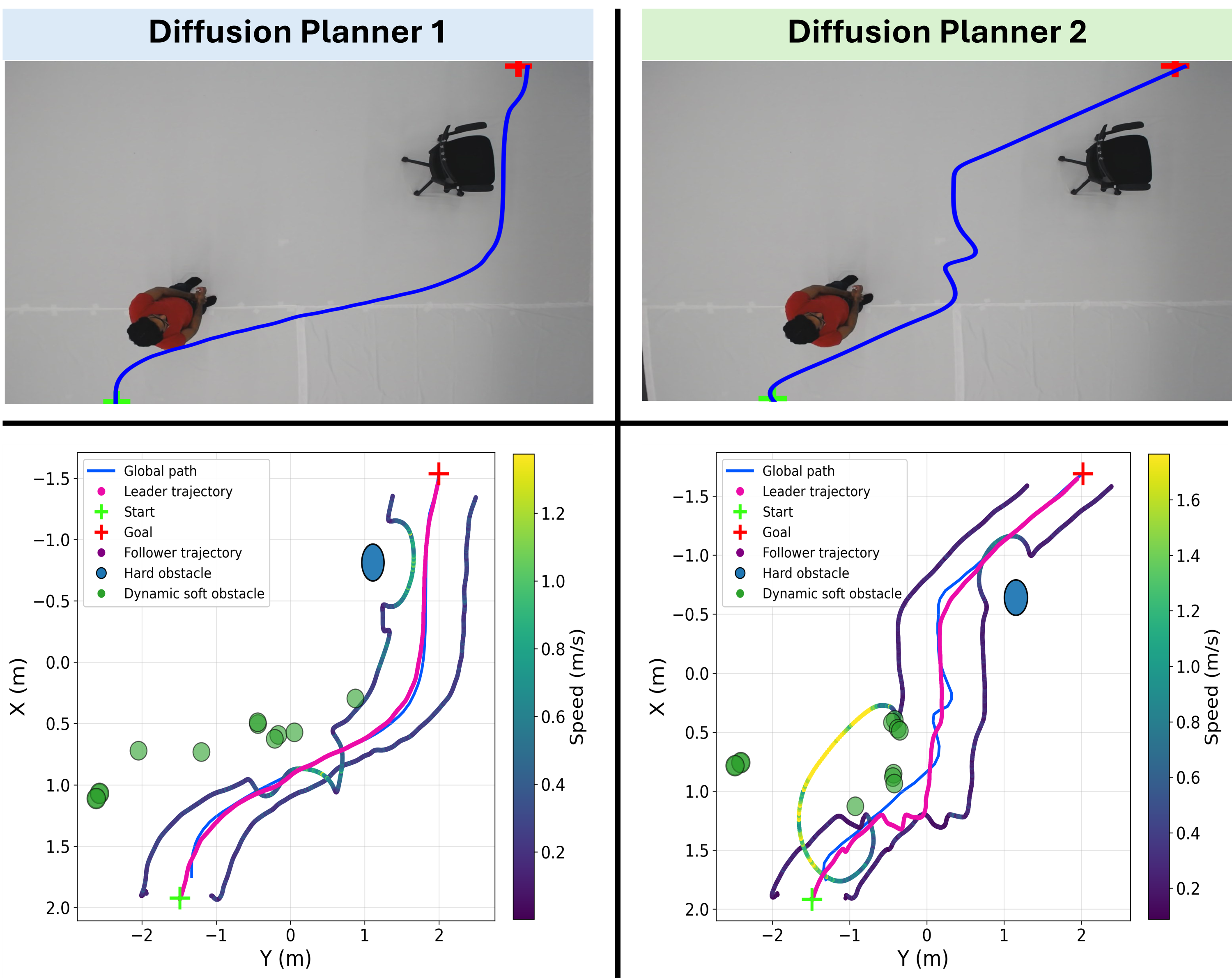}
    \caption{Dynamic environment with one moving human and one hard obstacle. Soft compliance near the human and rigid compliance near the chair are observed.}
    \label{fig:dynamic_exp_01}
\end{figure}

Fig.~\ref{fig:dynamic_exp_01} presents a dynamic environment containing a moving human and a static chair. The planners exhibit distinct behaviors. Planner 1 generates trajectories that pass closer to obstacles, whereas Planner 2 maintains larger local clearance due to FPV-based conditioning, leading to different velocity profiles.
Near the human, Planner 1 reduces speed to approximately 0.8 m/s, while Planner 2 maintains a higher speed of about 1.6 m/s. Higher velocity near a moving human reduces reaction margin and may increase risk in rapidly changing scenarios.

Rigid compliance is observed around the chair, characterized by small separation distances. In contrast, soft compliance near the human results in larger path deflections. Because Planner 1 operates closer to the human, it experiences stronger APF corrections and sharper local velocity variations.
\subsubsection{Experiment 4: Two Dynamic Soft Obstacles}

\begin{figure}[htbp]
    \centering
    \includegraphics[width=\columnwidth]{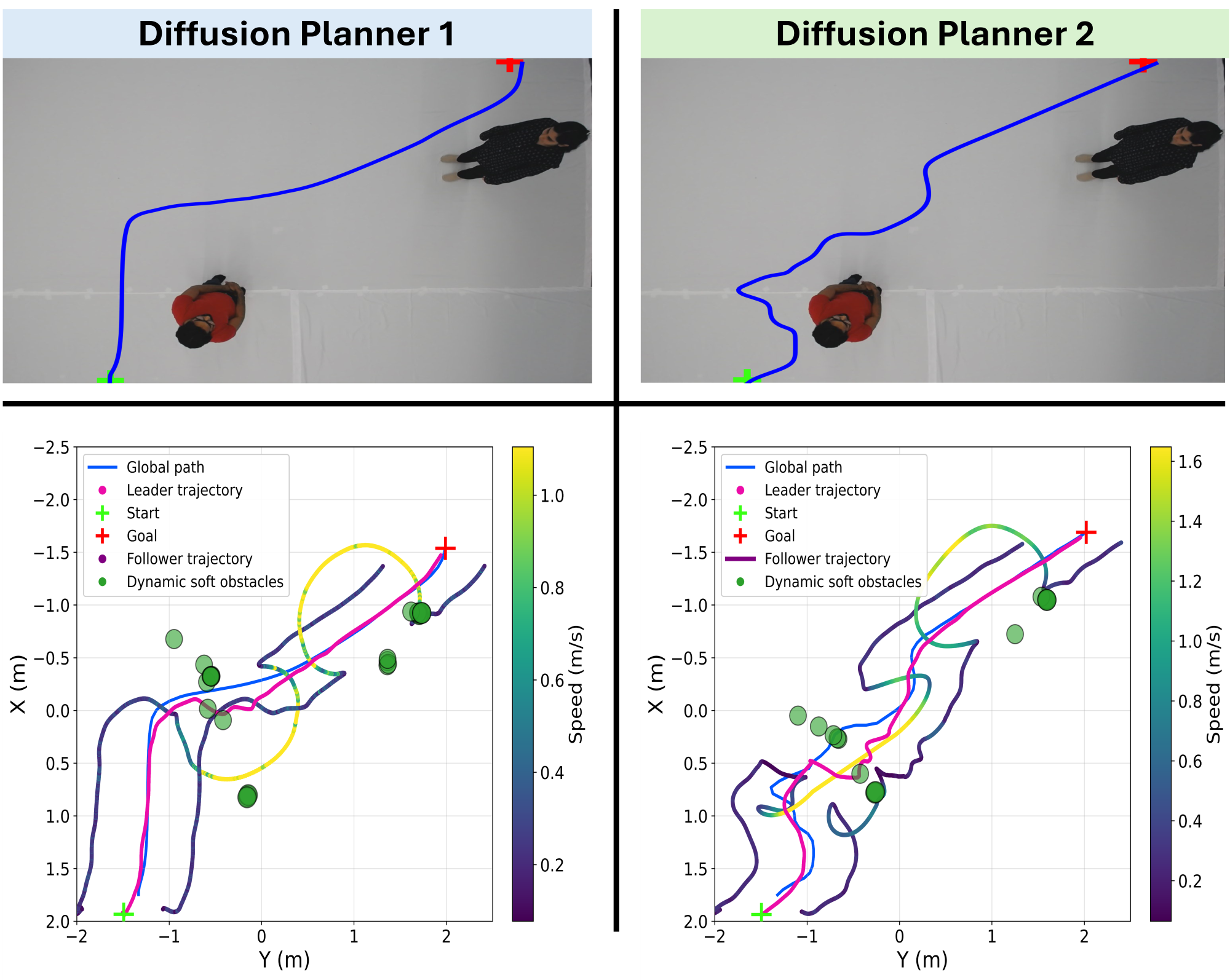}
    \caption{Dynamic scenario with two humans. Soft compliance with large deflection is observed.}
    \label{fig:dynamic_exp_02}
\end{figure}

Fig.~\ref{fig:dynamic_exp_02} presents a fully dynamic scenario with two moving humans. Both planners exhibit soft compliant behavior, maintaining large deflection distances around the humans. During avoidance, the follower speed is approximately 1.0 m/s for Planner 1 and 1.6 m/s for Planner 2.

Planner 2 sustains higher speeds due to its FPV-based local conditioning, which produces trajectories with greater immediate clearance and consequently smaller APF-induced repulsive forces. As a result, fewer corrective actions are required compared to Planner 1.
Minor trajectory overlaps and small jitters appear due to impedance-based leader–follower coupling. The oscillations remain bounded and decay over time, confirming stable closed-loop impedance behavior under dynamic conditions.

\subsection{Comparative Analysis of Diffusion Planners}
We compare two diffusion-based trajectory planners trained under different visual modalities. 
\textbf{Diffusion Planner 1 (P1)} is trained on top-view images and requires a single inference requiring 1.4--2.5\,s per trajectory. 
\textbf{Diffusion Planner 2 (P2)} is trained on first-person-view (FPV) images and deployed on top-view inputs using a two-stage inference pipeline, resulting in a longer inference time of 2.5--3.4\,s per trajectory.

Both planners achieved a \textbf{100\% trajectory generation rate} across all eight experimental scenarios. Since no trajectory generation failures were observed, the comparison focuses on trajectory quality metrics rather than binary success.

The evaluation metrics are:
\begin{itemize}
    \item \textbf{Path Length (m):} Total arc length of the trajectory.
    \item \textbf{Collision Ratio (-):} Fraction of trajectory samples intersecting inflated obstacle regions. High collision ratios correspond to overlap with inflated pixel-space safety regions and do not indicate physical collisions during real-world execution.
    \item \textbf{Goal Error (m):} Euclidean distance between the final trajectory point and the ground-truth goal.
    \item \textbf{Total Turning (rad):} Cumulative absolute heading change along the trajectory. Lower cumulative turning suggests smoother and more dynamically efficient trajectories.
\end{itemize}

\begin{table}[t]
\centering
\caption{\textsc{Performance Comparison Across Eight Experimental Scenarios}}
\footnotesize
\setlength{\tabcolsep}{3pt}
\begin{tabular}{@{}l l c c c c c@{}}
\hline
\textbf{Exp.} & \textbf{Scenario} & \textbf{Planner} & \textbf{Path} & \textbf{Coll.} & \textbf{Goal} & \textbf{Turn} \\
              &                   &                  & \textbf{(m)}  & \textbf{(-)}   & \textbf{(m)}  & \textbf{(rad)} \\
\hline
E1 & Cyl.--Gate       & P1 & 2.013 & 0.186 & 0.026 & 10.541 \\
   &                 & P2 & 2.191 & 0.067 & 0.028 & 7.279 \\
\hline
E2 & 2 Cyl.--Gate      & P1 & 2.148 & 0.352 & 0.044 & 9.254 \\
   &                 & P2 & 2.235 & 0.146 & 0.047 & 9.022 \\
\hline
E3 & Cyl.--Gate--Chair   & P1 & 2.093 & 0.338 & 0.016 & 11.333 \\
   &                 & P2 & 2.270 & 0.369 & 0.147 & 8.553 \\
\hline
E4 & Chair--Trolley     & P1 & 2.198 & 0.486 & 0.050 & 9.811 \\
   &                 & P2 & 2.191 & 0.295 & 0.050 & 10.216 \\
\hline
E5 & 5 Cyl.            & P1 & 2.105 & 0.693 & 0.044 & 9.190 \\
   &                 & P2 & 2.351 & 0.447 & 0.045 & 11.812 \\
\hline
E6 & 4 Cyl.--Human     & P1 & 2.078 & 0.380 & 0.044 & 9.893 \\
   &                 & P2 & 2.366 & 0.447 & 0.045 & 9.304 \\
\hline
E7 & Human--Chair       & P1 & 2.338 & 0.318 & 0.034 & 8.099 \\
   &                 & P2 & 2.354 & 0.119 & 0.034 & 15.313 \\
\hline
E8 & 2 Human          & P1 & 2.339 & 0.036 & 0.045 & 7.117 \\
   &                 & P2 & 2.509 & 0.085 & 0.047 & 16.760 \\
\hline
\end{tabular}
\end{table}

Across all scenarios, both planners produce comparable path lengths, confirming that the FPV-based planner can effectively perform global planning when required. The average path length difference between P1 and P2 remains marginal across experiments.

Planner~2 demonstrates a lower average collision ratio of (\textbf{0.246}) compared to Planner~1 (\textbf{0.348}), indicating improved obstacle clearance despite its cross-modal deployment, while goal accuracy remains comparable across all scenarios. In terms of geometric complexity, Planner~1 exhibits lower average cumulative turning (\textbf{9.41\,rad} vs.\ \textbf{11.03\,rad}), suggesting more directionally efficient trajectories, whereas the increased turning in Planner~2 is consistent with its iterative FPV-based inference strategy. Overall, these results indicate that Planner~1 offers faster inference and slightly smoother paths, while Planner~2 achieves competitive global performance with stronger clearance behavior. Across all experiments, the framework validates online impedance switching based on obstacle type, where hard obstacles induce small deflections due to higher stiffness and humans result in softer, larger compliant deviations, and impedance coupling preserves stable formation with only minor bounded oscillations and no collisions.

\section{Conclusion and Future Work}
This paper introduced \textit{ImpedanceDiffusion}, a hierarchical swarm navigation framework that leverages image-conditioned diffusion-based global planning, APF-based reactive execution, and semantic-aware variable impedance control. The proposed system enables map-free navigation of aerial drone swarms in mixed indoor environments with soft and hard obstacles without explicit geometric mapping or classical search-based planners.

Across 20 experimental configurations (100 total flight runs), the framework achieved a \textbf{92\% success rate}, with failures primarily attributed to hardware or communication limitations rather than planning instability. Both diffusion planners demonstrated a \textbf{100\% trajectory generation rate} across all evaluated scenarios, confirming the reliability of diffusion-based global planning under diverse environmental conditions. The VLM--RAG module achieved a \textbf{90\% retrieval accuracy} in identifying obstacle classes and selecting corresponding impedance parameters. Quantitatively, the FPV-conditioned planner exhibited stronger obstacle clearance behavior (lower average collision ratio of 0.246 compared to 0.348 for the top-view planner), while the top-view planner produced smoother trajectories with lower cumulative turning and faster inference. These results highlight a clear trade-off between geometric efficiency and local clearance robustness depending on visual conditioning modality. Overall, the findings demonstrate that diffusion models can serve as reliable global planners for image-conditioned swarm navigation, while semantic impedance adaptation enables obstacle-specific compliant behavior without compromising closed-loop stability. The integration of generative planning with physically grounded impedance control provides a practical pathway toward scalable, map-free, and semantically aware swarm navigation.

Future work will focus on enabling onboard perception and decentralized inference to remove reliance on workstation-level computation. Additionally, learning continuous impedance adaptation policies rather than discrete class-based switching will further improve scalability in dense, human-shared environments.

\balance

\bibliographystyle{IEEEtran}
\bibliography{bibliography}
\balance
\addtolength{\textheight}{-12cm}
\end{document}